\documentclass[10pt,twocolumn]{article}

\usepackage[utf8]{inputenc}
\usepackage[T1]{fontenc}
\usepackage{amsmath,amssymb}
\usepackage{graphicx}
\usepackage[margin=2cm]{geometry}
\usepackage{booktabs}
\usepackage{natbib}
\usepackage{hyperref}
\usepackage{xcolor}
\usepackage{caption}
\usepackage{float}

\hypersetup{colorlinks=true,linkcolor=blue!60!black,citecolor=blue!60!black,urlcolor=blue!60!black}

\title{%
  Supervised Training Rapidly Degrades Early Visual Cortex Alignment\\
  Across Biologically Plausible Learning Rules%
}

\author{%
  Nils Leutenegger\\
  Independent Researcher, Switzerland\\
  \texttt{github.com/nilsleut}%
}

\date{May 2026}

\newcommand{\newid}{arXiv:2608.12408}

\begin{document}
\maketitle

\begin{abstract}
\noindent\textbf{Correction (August 2026): the central finding of this paper is not
supported.} An evaluation-mode defect left the batch-normalisation layers of the
predictive-coding and STDP conditions in training mode during feature extraction,
producing their apparent preservation of V1 alignment. With the defect repaired,
predictive coding degrades V1 alignment \emph{more} than backpropagation does, not
less. The finding that training degrades V1 alignment for every rule tested does
survive. See the correction note following this abstract; the text below is
unchanged from v1.

\medskip

Random, untrained neural networks consistently match or exceed trained networks in representational similarity to early visual cortex. This puzzling finding challenges the assumption that learning improves brain alignment. We investigate it by tracking representational similarity analysis (RSA) alignment to human fMRI data across training for four learning rules: backpropagation (BP), feedback alignment (FA), predictive coding (PC), and spike-timing-dependent plasticity (STDP). Using 720 object images from the THINGS database and fMRI data from three subjects across six visual ROIs, we measure Spearman correlations between model and brain representational dissimilarity matrices at eight training checkpoints (epochs 0--40). We find that (1)~a single epoch of training reduces V1 alignment by 25--90\%, depending on the learning rule; (2)~backpropagation reduces V1 alignment most severely ($\Delta r = -0.080$), while predictive coding and STDP preserve substantially more ($\Delta r \approx -0.04$); and (3)~a weaker, opposite tendency appears in object-selective cortex (LOC), where BP shows the largest increase in alignment during training, although the absolute change is small. These results suggest that untrained architectures capture low-level visual statistics through inductive biases alone, and that global error signals (BP) reshape early representations more aggressively than local learning rules (PC, STDP), which better preserve brain-like structure.
\end{abstract}

\section*{Correction note (August 2026)}

\noindent The degradation of V1 alignment with training reported here is not robust to the analysis choices examined in ``Analysis choices decide whether untrained networks beat trained ones at V1: a multiverse analysis of model--brain RSA'' arXiv:2611.15533. In the model set re-tested there, with standardised model units, a fixed Conv1 layer and 224 px input, backpropagation after training scores above the untrained network at V1 (+0.011 [0.005, 0.017]). This covers the endpoint only; the trajectory and the best-layer analyses were not re-tested. The statement of the August note that the degradation of V1 alignment survives for every rule tested is likewise not robust to these choices at the endpoint. The text below is unchanged.

\noindent\textbf{The defect.} The predictive-coding and STDP model classes override
\texttt{eval()} with a no-op, so feature extraction left their batch-normalisation
layers in training mode and normalised every evaluation batch by its own statistics,
while the random, backpropagation and feedback-alignment conditions used stored
running statistics. Re-normalising each evaluation batch removes precisely the drift
in activation statistics that the degradation metric of this paper measures. The
apparent immunity of the local rules is what that defect produces, whether or not any
underlying effect exists.

\noindent\textbf{What the repaired analysis shows.} Re-running the full design with
the defect fixed leaves backpropagation and feedback alignment essentially unchanged
(change in the epoch $0 \to 40$ V1 difference of $+0.0001$ and $+0.0004$ at
$224$\,px). It reverses the result for the two affected conditions. Predictive coding
moves from $+0.0010 \pm 0.0005$ to $-0.0403 \pm 0.0034$ ($5/5$ seeds negative) and
STDP from $+0.0044 \pm 0.0034$ to $-0.0175 \pm 0.0033$ ($5/5$ seeds negative).
Predictive coding now degrades V1 alignment more than backpropagation does
($-0.040$ vs.\ $-0.031$), not less.

The claim that local learning rules preserve substantially more V1-like structure
than backpropagation --- the abstract, the three key findings, Sections~3.1--3.4 and
the discussion in Sections~4.1--4.3 --- is therefore not supported. The finding that
training degrades V1 alignment for every rule tested does survive, and is
strengthened.

\noindent\textbf{Two further qualifications.} First, in the implementation used here
predictive coding displaces its first convolutional layer by about $2\%$ of the
initialisation norm over $40$ epochs and does not update its third layer at all. A
network whose filters barely move is unlikely to change its alignment by $0.04$
through its weights; this is consistent with the trajectory reported for that
condition being carried by its accumulating normalisation statistics rather than by
its weight-update rule, though we have not tested that directly. It should in any
case not be read as a property of predictive coding as a learning algorithm. Second,
a companion study (\newid) shows that most of the apparent degradation at $224$\,px
in backpropagation depends on the evaluation resolution: evaluated at the $32$\,px
training resolution the same run returns to its starting value ($-0.000 \pm 0.005$,
$3/5$ seeds). That paper contains the corrected training-trajectory analysis, and
readers wanting the current state of this question should go there.

This paper reports best-layer-per-ROI alignment, whereas the corrected analysis uses
a fixed layer-to-ROI mapping, so absolute values are not directly comparable between
the two; the directions and the seed-level consistency are. The body of this version
is unchanged from v1 and should be read as superseded on the points above. The
repaired implementation and all re-run results are at
\url{https://github.com/nilsleut}.

\bigskip

\section{Introduction}

A growing body of work demonstrates that deep neural networks trained on visual tasks develop internal representations that correlate with neural responses in primate visual cortex \citep{yamins2014,khaligh2014,cichy2016}. This alignment has been taken as evidence that optimising for ecologically relevant tasks produces brain-like computations. However, a surprising counter-finding has emerged: untrained, randomly initialised networks often match or outperform trained networks in alignment with early visual areas, particularly V1 \citep{leutenegger2025}.

This raises a fundamental question: does learning improve brain alignment, or does it erode it? Prior work has compared learning rules at a single trained endpoint \citep{leutenegger2025,lillicrap2016}, but the dynamics of how alignment changes during training remain unexplored. Understanding these dynamics could reveal whether the random-weights advantage reflects an intrinsic property of network architectures or an active degradation caused by training.

We address this gap by measuring representational similarity to human fMRI across training for four learning rules spanning a spectrum of biological plausibility: backpropagation (BP), feedback alignment (FA) \citep{lillicrap2016}, predictive coding (PC) \citep{rao1999,whittington2017}, and spike-timing-dependent plasticity (STDP) \citep{bi1998,masquelier2007}. By extracting model representational dissimilarity matrices (RDMs) at eight checkpoints during 40 epochs of CIFAR-10 training and comparing them against fMRI RDMs from the THINGS dataset \citep{hebart2019,hebart2023}, we track how each learning rule reshapes representational geometry relative to the visual cortex.

Our results reveal three key findings. First, all learning rules degrade V1 alignment, but they do so at dramatically different rates: BP reduces V1 alignment by 90\% within a single epoch, while PC and STDP preserve approximately 70\%. Second, the magnitude of degradation tracks the globality of the error signal: rules that propagate precise, layer-specific error gradients (BP) are more destructive than rules relying on local computation (PC, STDP). Third, a weaker, opposite tendency appears in object-selective cortex (LOC), where BP shows the largest gain in alignment during training, suggesting that the same mechanism that erodes V1 structure may also build task-relevant representations in higher areas.

\section{Methods}

\subsection{Network Architecture}

All learning rules were implemented on a shared convolutional architecture consisting of three convolutional blocks (Conv1: 32 filters, Conv2: 64 filters, Conv3: 128 filters; each with $3 \times 3$ kernels, batch normalisation, ReLU, and $2 \times 2$ max-pooling) followed by a fully connected layer (FC1: 512 units) and a classification head (10 classes). This architecture was chosen to match \citet{leutenegger2025} and to ensure that differences between conditions reflect the learning rule, not the architecture.

\subsection{Learning Rules}

\textbf{Backpropagation (BP).} Standard supervised training with cross-entropy loss, Adam optimiser ($\text{lr} = 10^{-3}$, weight decay $10^{-4}$), cosine annealing schedule, gradient clipping at 1.0, and dropout (0.3).

\textbf{Feedback Alignment (FA).} Following \citet{lillicrap2016}, the backward pass uses fixed random feedback weights at all convolutional layers instead of the transpose of the forward weights. This replaces the symmetric weight transport required by BP with a biologically more plausible asymmetric pathway. SGD optimiser ($\text{lr} = 5 \times 10^{-4}$, momentum 0.9).

\textbf{Predictive Coding (PC).} Following \citet{rao1999,whittington2017}, each layer maintains a prediction of the layer below via learned transpose convolutions. During inference, representations are iteratively refined for $T = 10$ steps to minimise prediction errors. Feedforward weights are updated using local prediction error signals (learning rate $10^{-4}$), and a separate classifier head is trained with Adam.

\textbf{STDP.} Following \citet{masquelier2007}, convolutional weights are updated using spike-timing correlations: input activations are converted to Poisson spike trains, and weight changes follow an exponential STDP kernel ($A_+ = A_- = 0.003$, $\tau_+ = \tau_- = 20$ms, $T_{\text{sim}} = 10$ steps). A supervised classifier head is trained separately with Adam.

\textbf{Random Weights.} The untrained baseline uses the same architecture as BP (including batch normalisation and dropout) at initialisation (epoch~0).

\subsection{Training}

All models were trained on a random subset of 8{,}000 CIFAR-10 training images (batch size 128) for 40 epochs. Five random seeds (42, 123, 456, 789, 1337) were used per learning rule. Model activations were extracted at eight checkpoints: epochs 0, 1, 2, 5, 10, 20, 30, and 40. Epoch~0 corresponds to the untrained random-weight baseline.

\subsection{fMRI Data}

We use the THINGS-fMRI dataset \citep{hebart2023}, which provides blood-oxygen-level-dependent (BOLD) responses from three human subjects viewing naturalistic object images. We selected 720 images for which fMRI data were available across all subjects. Responses were extracted from six regions of interest (ROIs): V1, V2, V3, V4, LOC, and IT. Subject-level representational dissimilarity matrices (RDMs) were computed using correlation distance.

\subsection{Representational Similarity Analysis}

At each checkpoint, 720 THINGS images ($224 \times 224$ pixels, ImageNet normalisation) were passed through the model. Layer activations were global-average-pooled to produce feature vectors, and model RDMs were computed using correlation distance. Brain--model alignment was quantified as the Spearman rank correlation between the upper triangles of the model and fMRI RDMs. We report the best-layer alignment per ROI (i.e., the layer yielding the highest Spearman $r$ for each ROI, evaluated independently at each epoch).

\subsection{Statistical Testing}

All statistical comparisons use paired, one-sided permutation tests across the five seeds, with the seed serving as the pairing unit. With five seeds there are only $2^5 = 32$ possible sign assignments, so the smallest attainable one-sided $p$-value is $1/32 \approx 0.031$; a comparison reaches this floor exactly when the effect is consistent in direction across all five seeds. We therefore report $p = 0.031$ as evidence of a fully consistent directional effect rather than of a small tail probability, and we treat the five-seed design as a limit on statistical resolution (Section~\ref{sec:limitations}). Significance was assessed at $\alpha = 0.05$. Cohen's $d$ effect sizes are reported for key comparisons.

\section{Results}

\subsection{Training Universally Degrades V1 Alignment}

\begin{figure*}[t]
  \centering
  \includegraphics[width=\textwidth]{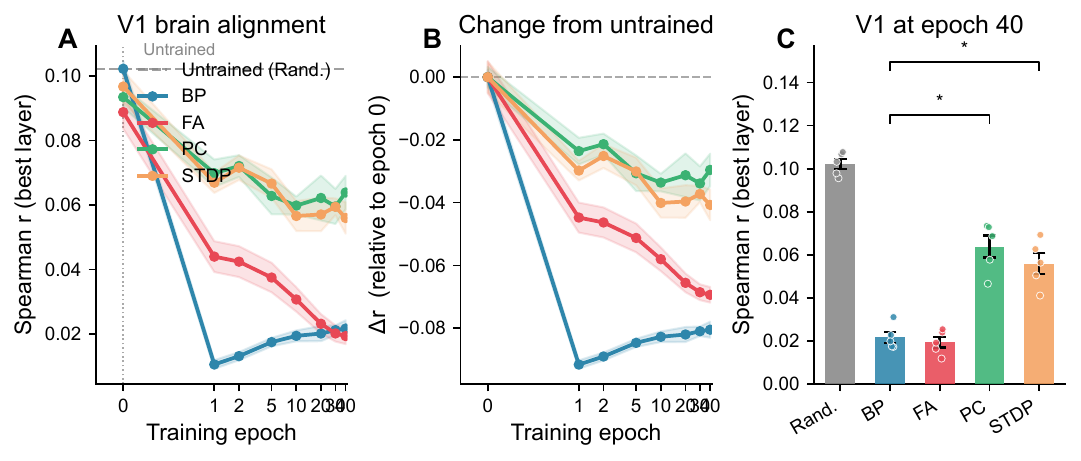}
  \caption{\textbf{V1 brain alignment across training.}
  \textbf{(A)}~Spearman $r$ between model RDMs (best layer) and V1 fMRI RDMs across training epochs. Shaded regions: $\pm$1 SD across 5 seeds. Dashed grey line: untrained baseline.
  \textbf{(B)}~Change in alignment relative to epoch~0. BP (blue) drops most steeply; PC (green) and STDP (orange) degrade least.
  \textbf{(C)}~Final alignment at epoch~40. Stars indicate significant difference from untrained baseline (paired permutation test, $p < 0.05$). Individual seed values shown as dots.}
  \label{fig:v1}
\end{figure*}

At initialisation (epoch~0), all models show comparable V1 alignment (Spearman $r \approx 0.09$--$0.10$; Figure~\ref{fig:v1}A). After a single epoch of training, all learning rules show reduced V1 alignment, but the magnitude differs dramatically across rules (Figure~\ref{fig:v1}B). Backpropagation shows the most severe degradation, losing 90\% of its V1 alignment after one epoch ($r: 0.102 \to 0.011$, paired permutation test $p = 0.031$). Feedback alignment shows an intermediate drop of 49\% ($r: 0.089 \to 0.044$). Predictive coding and STDP show the least degradation, losing only 25\% and 31\% respectively (PC: $r: 0.093 \to 0.070$; STDP: $r: 0.097 \to 0.067$).

By epoch~40, the ordering stabilises at PC ($r = 0.064 \pm 0.012$) $>$ STDP ($0.059 \pm 0.010$) $>$ BP ($0.022 \pm 0.006$) $\approx$ FA ($0.019 \pm 0.006$). Both PC and STDP retain significantly more V1 alignment than BP (paired permutation test, $p = 0.031$, the resolution floor for five seeds; Cohen's $d > 5$ for both, reflecting the very low between-seed variance). All trained models show significantly lower V1 alignment than the untrained baseline ($p = 0.031$ for all comparisons, that is, every seed showed the same ordering; Figure~\ref{fig:v1}C).

\subsection{The Degradation Pattern Generalises Across Early Visual Areas}

\begin{figure*}[t]
  \centering
  \includegraphics[width=\textwidth]{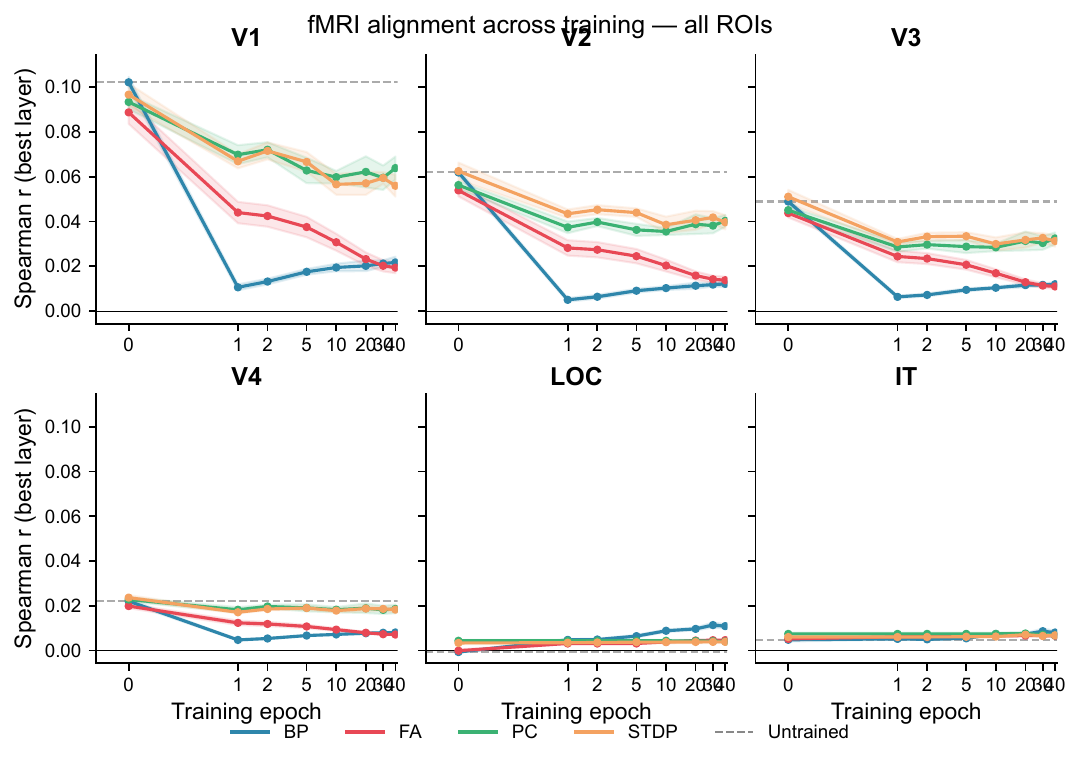}
  \caption{\textbf{fMRI alignment across training for all six ROIs.}
  Same conventions as Figure~\ref{fig:v1}A. The degradation pattern is strongest in early visual areas (V1--V3) and absent in higher areas (LOC, IT).}
  \label{fig:allrois}
\end{figure*}

The V1 degradation pattern extends to V2 and V3, with BP consistently showing the largest drop and PC/STDP preserving the most alignment (Figure~\ref{fig:allrois}). In V4, the pattern attenuates: all rules show moderate degradation with smaller differences between them. Notably, in LOC and IT, the trends reverse or flatten (see Section~\ref{sec:loc}).

\subsection{Opposing Trend in Object-Selective Cortex}
\label{sec:loc}

\begin{figure*}[t]
  \centering
  \includegraphics[width=\textwidth]{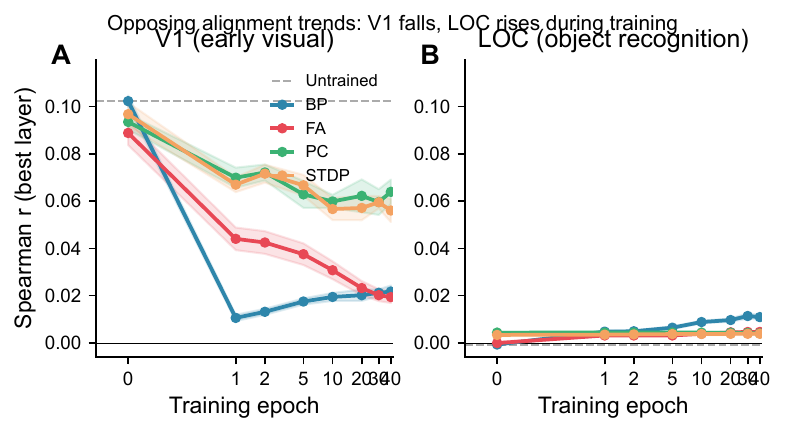}
  \caption{\textbf{Opposing trends in V1 and LOC.}
  \textbf{(A)}~V1 alignment decreases during training for all rules.
  \textbf{(B)}~LOC alignment increases for BP only, while local rules (PC, STDP) show no change. Same y-axis scale highlights the magnitude difference.}
  \label{fig:v1loc}
\end{figure*}

While training degrades alignment with early visual cortex, the opposite tendency appears, weakly, in LOC (Figure~\ref{fig:v1loc}). Backpropagation, the rule that degrades V1 alignment most severely, shows the largest increase in LOC alignment during training (epoch~0: $r = -0.001$; epoch~40: $r = 0.011$; $\Delta r = +0.011$). The other rules show smaller or negligible changes (FA: $\Delta r = +0.005$; PC: $\Delta r = -0.001$; STDP: $\Delta r = +0.001$). These LOC changes are small in absolute terms and we did not subject them to a significance test, so they should be read as a suggestive tendency rather than an established effect.

This dissociation suggests a trade-off: BP's global error signal reshapes representations throughout the network, degrading early visual structure while building task-relevant object representations in higher layers. Local learning rules, by contrast, lack the top-down pressure needed to sculpt higher-layer representations but consequently preserve early visual statistics.

\subsection{Seed Variability}

\begin{figure*}[t]
  \centering
  \includegraphics[width=0.8\textwidth]{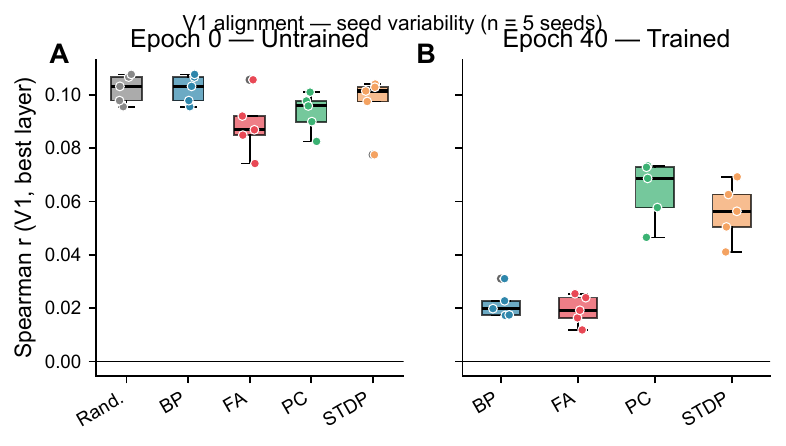}
  \caption{\textbf{Seed variability.}
  \textbf{(A)}~V1 alignment at epoch~0 (untrained). All rules cluster near $r \approx 0.10$.
  \textbf{(B)}~V1 alignment at epoch~40 (trained). BP and FA collapse to $r \approx 0.02$; PC and STDP retain $r \approx 0.06$. Box plots with individual seed values (circles).}
  \label{fig:seeds}
\end{figure*}

At epoch~0, all architectures cluster tightly around $r \approx 0.09$--$0.10$ with low seed variability (Figure~\ref{fig:seeds}A). After training, the rules separate into two groups: BP and FA converge to low alignment ($r \approx 0.02$) with low variance, while PC and STDP maintain higher alignment ($r \approx 0.06$) with moderate variance (Figure~\ref{fig:seeds}B). The consistency across seeds confirms that the observed differences reflect systematic properties of the learning rules rather than random fluctuations.

\section{Discussion}

\subsection{Why Does Training Degrade V1 Alignment?}

Our central finding, that supervised training degrades V1 alignment across all learning rules, suggests that untrained networks capture low-level visual statistics through their architectural inductive biases (convolutional filters, pooling, normalisation) rather than through learning. Training then reshapes these representations toward task-relevant features, moving them away from the general-purpose visual statistics encoded by V1. This view is consistent with work showing that explicitly aligning a network's early layers to primate V1 reshapes its representations and improves robustness \citep{safarani2021}, which indicates that V1-like structure is a specific, shapeable property rather than a generic by-product of training.

The key insight is that the \emph{degree} of degradation depends on the learning rule. BP, which computes exact gradients and propagates precise error signals across all layers, reshapes representations most aggressively. FA, which substitutes random feedback weights, delivers noisier error signals and degrades V1 alignment less rapidly (though it converges to similar levels by epoch~40). PC and STDP, which rely on local computation without top-down error propagation, preserve substantially more V1-like structure throughout training.

\subsection{A Trade-Off Between Early and Higher-Level Alignment}

The opposing trends in V1 (degradation) and LOC (a small increase for BP) suggest a representational trade-off. BP's global gradient signal may simultaneously degrade early visual structure and build category-selective representations. Local learning rules avoid this trade-off: they preserve V1 alignment but fail to develop LOC-aligned representations, suggesting that local credit assignment is insufficient for building hierarchical object representations.

This finding has implications for theories of cortical learning. The brain appears to maintain strong V1 representations while also developing object selectivity in LOC and IT, a combination that none of our tested rules achieves. This may point to a learning mechanism that combines local representational preservation with a form of hierarchical credit assignment, potentially more nuanced than any single rule tested here.

\subsection{Relation to Prior Work}

Our results complement and extend \citet{leutenegger2025}, which compared trained learning rules at a single endpoint and found that random weights outperform all trained models at V1. The present work reveals the \emph{dynamics} of this phenomenon: the random-weights advantage is not merely a static comparison but reflects an active, rapid degradation of V1-aligned structure by training. The finding that PC and STDP partially preserve V1 alignment is novel and connects to a broader literature on the biological plausibility of learning rules \citep{whittington2017,payeur2021}.

\subsection{Limitations}
\label{sec:limitations}

Several limitations warrant consideration. First, all rules share a common simple architecture (3~conv + 1~FC). Deeper architectures (e.g., ResNets) may show different dynamics. Second, the training dataset (CIFAR-10, 8{,}000 images) is small relative to natural vision, and the networks are trained on $32 \times 32$ CIFAR-10 images but evaluated on $224 \times 224$ THINGS images; this resolution and domain shift could itself affect the extracted representations, so the absolute alignment values should be interpreted with care. Third, the STDP and PC implementations are simplified approximations of their biological counterparts. Fourth, the fMRI dataset comprises only three subjects, and the five-seed design caps the resolution of the permutation tests at $p \approx 0.031$, limiting statistical power for subject- and seed-level analyses. Fifth, the LOC increases were small in absolute terms and were not tested for significance, so the proposed early-versus-higher trade-off should be regarded as suggestive rather than established. Finally, the effect sizes, while large in relative terms, operate on small absolute Spearman correlations ($r < 0.10$), so the practical significance of these representational differences for downstream neural processing remains an open question.

\subsection{Conclusion}

Supervised training universally degrades early visual cortex alignment, but the magnitude depends systematically on the learning rule. Backpropagation, the least biologically plausible rule, is the most destructive, while local learning rules (predictive coding, STDP) preserve substantially more V1-like structure. This pattern reveals a fundamental tension between task optimisation and representational brain-likeness that any theory of cortical learning must resolve.

\section*{Data and Code Availability}

All code is available at \url{https://github.com/nilsleut}. The THINGS-fMRI dataset is publicly available \citep{hebart2023}. Training dynamics data (model RDMs, checkpoints, and RSA results at all milestones) will be released upon publication.

\section*{Acknowledgements}

This work was conducted independently. Compute was provided by Modal (GPU cloud) and Kaggle Notebooks.

\bibliographystyle{plainnat}

\end{document}